\documentclass[10pt,twocolumn,letterpaper]{article}

\usepackage[pagenumbers]{cvpr} 
\usepackage{marvosym}
\usepackage{tabularx}
\usepackage{color}
\usepackage{colortbl}
\usepackage{xcolor}
\usepackage{multirow}
\usepackage{multicol}
\usepackage{wrapfig}

\definecolor{cvprblue}{rgb}{0.21,0.49,0.74}
\usepackage[pagebackref,breaklinks,colorlinks,citecolor=cvprblue]{hyperref}

\def\paperID{*****} 
\def\confName{CVPR}
\def\confYear{2024}

\title{\raisebox{-0.16cm}{\includegraphics[scale=1.1]{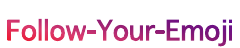}}:
Fine-Controllable and Expressive Freestyle Portrait Animation}

\author{Yue Ma$^{1}$\footnotemark[2], \quad Hongyu Liu$^{1}$\footnotemark[2],\quad  Hongfa Wang$^{2,3}$\footnotemark[2], \quad Heng Pan$^{2}$\footnotemark[2]\\
  Yingqing He$^{1}$, \quad  Junkun Yuan$^{2}$,\quad  Ailing Zeng$^{2}$, \quad  Chengfei Cai$^{2}$, \quad   Heung-Yeung Shum$^{1,3}$, \\ Wei Liu$^{2\textrm{\Letter}}$, \quad  Qifeng Chen$^{1\textrm{\Letter}}$ \\
  $^{1}$HKUST \quad  $^{2}$Tencent, Hunyuan \quad 
  $^{3}$Tsinghua University\\
\url{https://follow-your-emoji.github.io/}
}

\begin{document}

\renewcommand{\thefootnote}{\fnsymbol{footnote}}
\twocolumn[{
\maketitle
\begin{center}
    \captionsetup{type=figure}
    \includegraphics[width=1\linewidth]{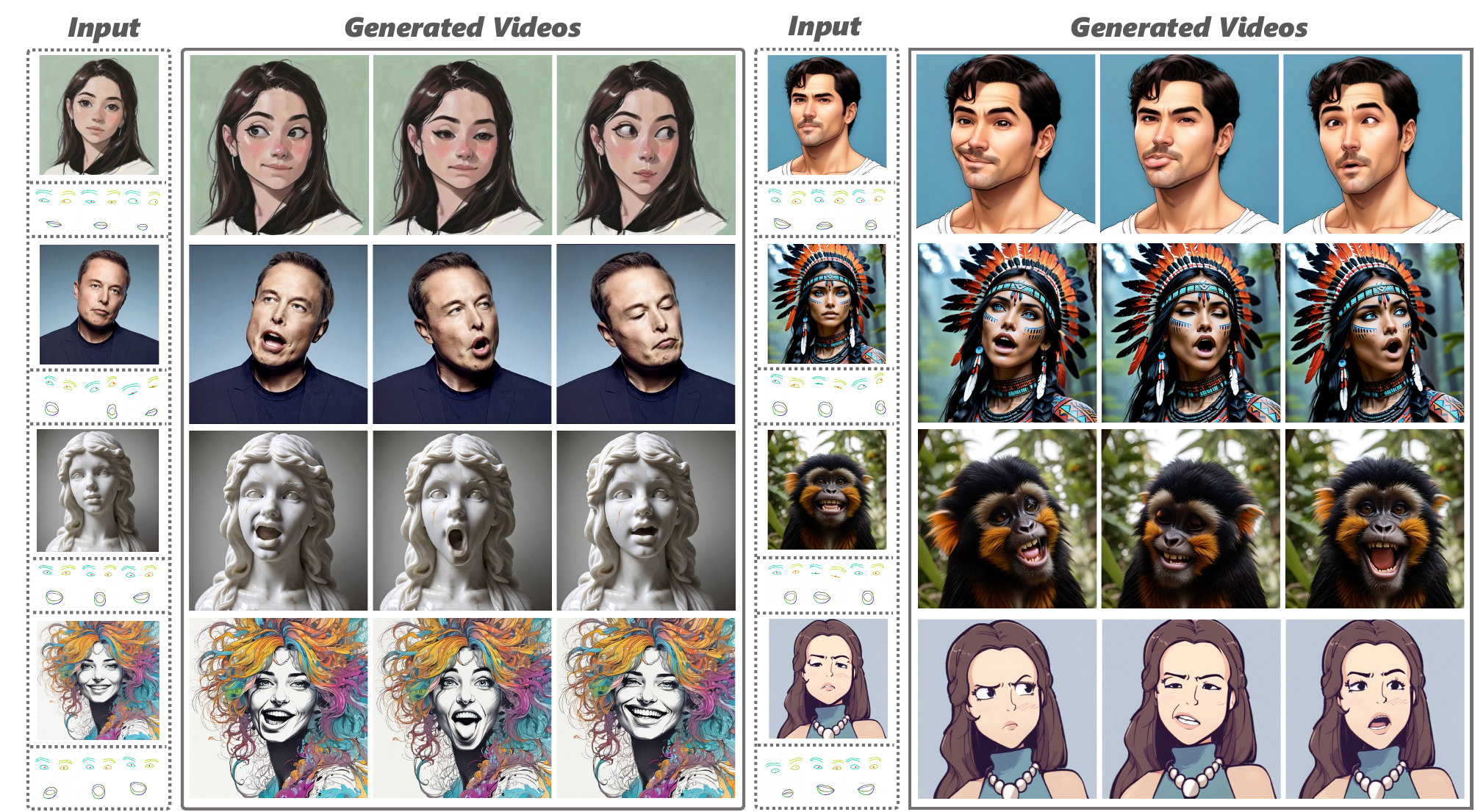}
    \captionof{figure}{\textbf{Qualitative results of our Follow-Your-Emoji.} The images of the input column are the reference portrait and the corresponding motion landmarks. 
  Using exaggerated expressions with landmark sequences, our portrait animation framework can animate freestyle reference portraits, e.g., cartoons, realism, sculptures, and even animals.}
\end{center}
\label{fig:teaser}
}]

\footnotetext[2]{Equal contribution.} \footnotetext[0]{${\textrm{\Letter}}$ Corresponding author.}


\begin{abstract}
We present Follow-Your-Emoji, a diffusion-based framework for portrait animation, which animates a reference portrait with target landmark sequences. The main challenge of portrait animation is to preserve the identity of the reference portrait and transfer the target expression to this portrait while maintaining temporal consistency and fidelity. To address these challenges, Follow-Your-Emoji equipped the powerful Stable Diffusion model with two well-designed technologies. Specifically,  we first adopt a new explicit motion signal, namely expression-aware landmark,  to guide the animation process. We discover this landmark can not only ensure the accurate motion alignment between the reference portrait and target motion during inference but also increase the ability to portray exaggerated expressions (i.e., large pupil movements) and avoid identity leakage. Then, we propose a facial fine-grained loss to improve the model's ability of subtle expression perception and reference portrait appearance reconstruction by using both expression and facial masks. 
Accordingly, our method demonstrates significant performance in controlling the expression of freestyle portraits, including real humans, cartoons, sculptures, and even animals. By leveraging a simple and effective progressive generation strategy, we extend our model to stable long-term animation, thus increasing its potential application value. To address the lack of a benchmark for this field, we introduce \textbf{EmojiBench}, a comprehensive benchmark comprising diverse portrait images, driving videos, and landmarks. We show extensive evaluations on EmojiBench to verify the superiority of Follow-Your-Emoji.

\end{abstract}    
\section{introduction}
We study the task of portrait animation, which transfers the target sequences of poses and expressions from the driven video to the reference portrait. Combined with the generative adversarial network~\cite{goodfellow2020generative} (GAN) and diffusion model~\cite{sohl2015deep}, recent portrait animation methods demonstrate widespread potential applications, such as online conferencing, virtual characters, and augmented reality.

For the GAN-based portrait animation method~\cite{megaPortriat, wang2021facevid2vid, fomm, liu2023human}, they typically utilize a two-stage pipeline which first warps the reference image in feature space with flow field, then adopts the GAN as a rendering decoder to refine the warping features and generate the missing or occluded body parts. However, due to the limited performance of GAN and the inaccuracy of motion representation of the flow field, the generation results of these methods always suffer from unrealistic content and remarkable artifacts. In recent years, diffusion models~\cite{ho2020denoising, song2020denoising} have showcased better generation ability than GAN. Some methods bring powerful foundation diffusion models for high-quality video~\cite{guo2023animatediff,blattmann2023align,ho2022imagen,ho2022video,wu2023tune,he2022lvdm,chen2024videocrafter2,chen2023videocrafter1} and image generation~\cite{rombach2021highresolution,saharia2022photorealistic,ramesh2022hierarchical, li2023finedance} with large-scale image or video datasets. However, these foundation models can not directly handle the main challenges of the portrait animation task: preserving the reference portrait's identity during animation and effectively modeling the target expression for the portrait.

Intuitively,  some methods~\cite{magicpose, xu2023magicanimate, zhu2024champ, hu2023animate, xu2023magicanimate, he2023camouflaged, hu2023animate, wang2023disco} try to modify the architecture of foundation diffusion model (i.e., Stable Diffusion~\cite{rombach2021highresolution}) with some plug and play modules for portrait animation task and leverage the pretrained diffusion model as powerful prior information. Specifically, they utilize an appearance net~\cite{hu2023animate} and CLIP model~\cite{radford2021learning} to extract identity information of the reference portrait and temporal attention to establish temporal consistency between frames. However, the video results of these methods exhibit distortions and unrealistic artifacts, especially when animating uncommon domain portraits (i.e., cartoons, sculptures, and animals) that are not represented in the training data. We find this is mainly due to two reasons:  (1) The motion representation (i.e., 2D landmarks~\cite{magicpose, hu2023animate} or the motion image itself~\cite{xie2024x}) adopted in these methods are not robust enough. During inference, 2D landmarks can easily lead to a misalignment between the facial features of the reference portrait and the target motion, resulting in identity leakage. However, setting the motion image itself as the signal needs to utilize third-party methods to change the identity of the target motion videos for training, as mentioned in Xportrait~\cite{xie2024x}. And it will destroy the subtle expression features in the original motion videos. (2) These methods utilize the original loss in the diffusion model during training, which is unsuitable for portrait animation tasks that need the model to focus on capturing reference facial appearance and expression changes.

In this paper, we present Follow-Your-Emoji, a novel diffusion-based framework for portrait animation. Apart from the commonly used appearance net and temporal attention in recent diffusion-based portrait animation methods, we propose several effective technologies to address the aforementioned problems.  (1) We introduce the expression-aware landmark, a novel expression control signal, to guide the driving process more effectively. Specifically, we obtain the landmark by projecting the 3D keypoints obtained from MediaPipe~\cite{mediapipe}. Owing to the inherent canonical property of 3D keypoints, we can effectively align the target motion with the reference portrait during inference, thereby avoiding identity leakage. However, MediaPipe is not robust enough, as the facial contour sometimes fails to conform to the face accurately. Consequently, the process of projecting landmarks has been modified to exclude facial contours and incorporate pupil points. This operation enables the model to better focus on expression changes (i.e., pupil point motion) while preventing it from influencing the shape and destroying the identity information of the reference portrait through the wrong facial contour.  (2) We propose a facial fine-grained loss function to aid the model in focusing on capturing subtle expression changes and the detailed appearance of the reference portrait.  Specifically, we first leverage both facial masks and expression masks with our expression-aware landmark, then compute the spatial distance between the ground truth and predicted results in these mask regions.  

Through the aforementioned improvements, our approach can effectively drive freestyle portraits, as illustrated in Figure~\ref{fig:teaser}. 
Additionally, to train our model, we construct a high-quality expression training dataset with 18 exaggerated expressions and 20-minute real-human videos from 115 subjects.
We employ a progressive generation strategy that enables our method to scale to long-term animation synthesis with high fidelity and stability. To address the lack of a benchmark in portrait animation, we introduce a comprehensive benchmark called EmojiBench, which consists of 410 various style portrait animation videos that showcase a wide range of facial expressions and head poses. 
Finally, we conduct a comprehensive evaluation of Follow-Your-Emoji using EmojiBench.
The evaluation results demonstrate the impressive performance of our method in handling portraits and motions that were outside of the training domain. Compared with the existing baseline methods, our method performs quantitatively and qualitatively better, delivering exceptional visual fidelity, faithful representation of identities, and precise motion rendering. In summary, our contributions can be summarized as follows:

\begin{itemize}
\item We introduce \textit{Follow-Your-Emoji}, a diffusion-based framework for fine-controllable portrait animation. Based on the proposed progressive generation strategy, it can further produce long-term animation.

\item To facilitate freestyle portrait animation, we propose the expression-aware landmarks as the motion representation and a facial fine-grinned loss to help the diffusion model enhance the generation quality of facial expressions.

\item To train our model, we introduce a new expression training dataset with 18 expressions and 20-min talking videos from 115 subjects. To validate the effectiveness of our methods, we construct a benchmark EmojiBench, and comprehensive results show the superiority of our Follow-Your-Emoji in fine-controllable and expressive aspects.

\end{itemize}

\section{Related Work}

\subsection{GAN-based Portrait Animation}

Animating a single portrait has attracted a lot of attention in the research. Previous approaches~\cite{megaPortriat, fomm, magicpose} mainly leverage Generative Adversarial Networks (GANs)~\cite{goodfellow2020generative} to generate plausible motion using self-supervised learning.  
The pioneering works primarily involved two steps: warping and rending. These methods firstly estimate head and facial motion with open-source 2D/3D pose predictors~\cite{mediapipe, dwpose}. The facial representation is warped and fed into a generative model to synthesize dynamic frames with realistic animation and rich details.
Following such a paradigm, a majority of approaches~\cite{wang2021facevid2vid, hong2022depth, TPS, ReenacArtFace} focus on improve facial warping estimation, including 3D neural landmarks~\cite{wang2021facevid2vid}, thin-plate splines~\cite{TPS} and depth~\cite{hong2022depth}. 
Additionally, the 3D morphable is utilized to model the expression and motion in ReenacArtFace~\cite{ReenacArtFace}. ToonTalker~\cite{gong2023toontalker} employs the transformer architecture to help the warping process of cross-domain datasets. 
MegaPortraits~\cite{megaPortriat} enhances rendered image quality using high-resolution image data, whereas FADM~\cite{zeng2023face} enriches generated details using the proposed coarse-to-fine animation framework. Face Vid2Vid~\cite{wang2021facevid2vid} presents a pure neural rendering to decompose identity-specific and motion-related information unsupervisedly.
In addition to video reenactment, there are also various driving signals, such as 3D facial prior~\cite{deng2020disentangled, feng2021learning, khakhulin2022realistic, sun2023next3d, xu2023omniavatar} and audio~\cite{tian2024emo, xu2024vasa, he2023gaia, zhang2023sadtalker, wei2024aniportrait}. However, these methods primarily focus on talking scenarios, and they struggle to synthesize animated frames with high-quality facial details and diverse domain styles. 


\subsection{Diffusion-based Portrait Animation}

Diffusion models (DMs)~\cite{ho2020denoising, song2020denoising} 
achieves superior performance in various generative tasks including image generation~\cite{rombach2021highresolution, zhao2019image, ruiz2023dreambooth, shi2024motion} and editing~\cite{cao2023masactrl, hertz2022prompt, brooks2023instructpix2pix}, video generation~\cite{ma2024follow, singer2022make, he2022latent, ma2024followyourclick, wang2024animatelcm} and editing~\cite{qi2023fatezero, zhang2023controlvideo, liu2023video, ma2023magicstick, li2024lodge}. Recently, latent diffusion models further improved the performance by operating the diffusion step in latent space. 
Mainstream portrait animation approaches leverage the power of Stable Diffusion (SD)~\cite{rombach2021highresolution} and incorporate temporal information into generation process, such as AnimateDiff~\cite{guo2023animatediff}, MagicVideo~\cite{zhou2022magicvideo}, VideoCrafter~\cite{chen2023videocrafter1} and ModelScope~\cite{wang2023modelscope}.
Additionally, current works~\cite{magicpose, xu2023magicanimate, he2023strategic, hu2023animate, zhu2024champ, gao2024inducing} employ the self-attention blocks with injected reference image to achieve identity preservation.  They always product high-quality video clips with textual guidance, which is ambiguous and struggle to describe the intention from users. To achieve more controllable generation, many signals are applied for video generation, such as depth map~\cite{gen2, xing2024make}, skeleton~\cite{magicpose, ma2024follow} and sketch~\cite{zhang2023controlvideo}. Another state-of-the-art works~\cite{magicpose, xu2023magicanimate, zhu2024champ, hu2023animate, gao2023backdoor} integrate the appearance and pose condition into temporal layers for full-body video generation.
However, these methods all focus on full-body animation and ignore the specific details of the face. In contrast, we innovate the diffusion-based framework, focusing on driving various style portraits with detailed facial expressions (\textit{e.g.}, eyes, skins). 


\section{Preliminaries}

\subsection{Latent Diffusion Model} 
Latent diffusion models (LDM)~\cite{rombach2021highresolution}, the most critical component of Stable Diffusion (SD), is a text-to-image diffusion model that reformulates the diffusion and denoising procedures within a latent space instead of image space for stable and fast training.  
The VAE~\cite{kingma2013auto} projects images from RGB space to latent space, where the diffusion process is guided by textual embedding.
Then, 
a UNet-based network~\cite{ronneberger2015u} incorporates self-attention and cross-attention mechanisms through Transformer Blocks to learn the reverse denoising process in latent space. The cross-attention helps the text prompt inject into the whole process in an effective manner. The whole training objective of the UNet can be written as: 
 

\begin{equation}
\mathcal{L}_{LDM}=\underset{t, z_, \epsilon}{\mathbb{E}} [ \|\epsilon-\epsilon_\theta (\sqrt{\bar{\alpha}_t} z +\sqrt{1-\bar{\alpha}_t} \epsilon, c, t ) \|^2 ]
\label{eq:ldm}
\end{equation}
where $z$ notes the latent embedding of training sample. $\epsilon_\theta$ and $\epsilon$ represent predicted noise by diffusion model and ground truth noise at corresponding timestep $t$, respectively. $c$ is the condition embedding involved in the generation and the coefficient $\bar{\alpha}_t$ remains consistent with that employed in vanilla diffusion models.


\subsection{Portrait Animation with Diffusion}
Recent methods~\cite{magicpose, xu2023magicanimate, zhu2024champ, hu2023animate, xie2024x} try to expand SD for full body or portrait animation. To facilitate the utilization of powerful pre-trained SD models, their frameworks exhibit substantial similarities, consisting of several plug-and-play modules. There are three main modules: (1) \noindent\textbf{Appearance Net: } It extracts the identity attributes and background context from the reference portrait first and then injects this information into UNet in SD by adding features to the self-attention blocks. The architecture of the appearance net is the same as the UNet in SD. (2) \noindent\textbf{Temporal Attention: } Equipped the UNet with temporal transformers to maintain the cross-frame correspondence and temporal coherence. (3) \noindent\textbf{Control Motion Injection:} To build the spatial mapping between the control signals and the output, these methods always utilize the ControlNet~\cite{zhang2023adding} or add the feature of motions to the input of UNet directly~\cite{hu2023animate}. (4) \noindent\textbf{Image Prompt Injection:} To transfer the UNet from text-to-image generation to portrait animation, the image prompt injection module replaces the text encoder of CLIP with the correspondence image encoder to get the token of the reference portrait image. Then, these tokens are sent to UNet with the cross-attention layer similar to the original text token in SD.


\begin{figure*}
  \centering
  \includegraphics[width=\textwidth]{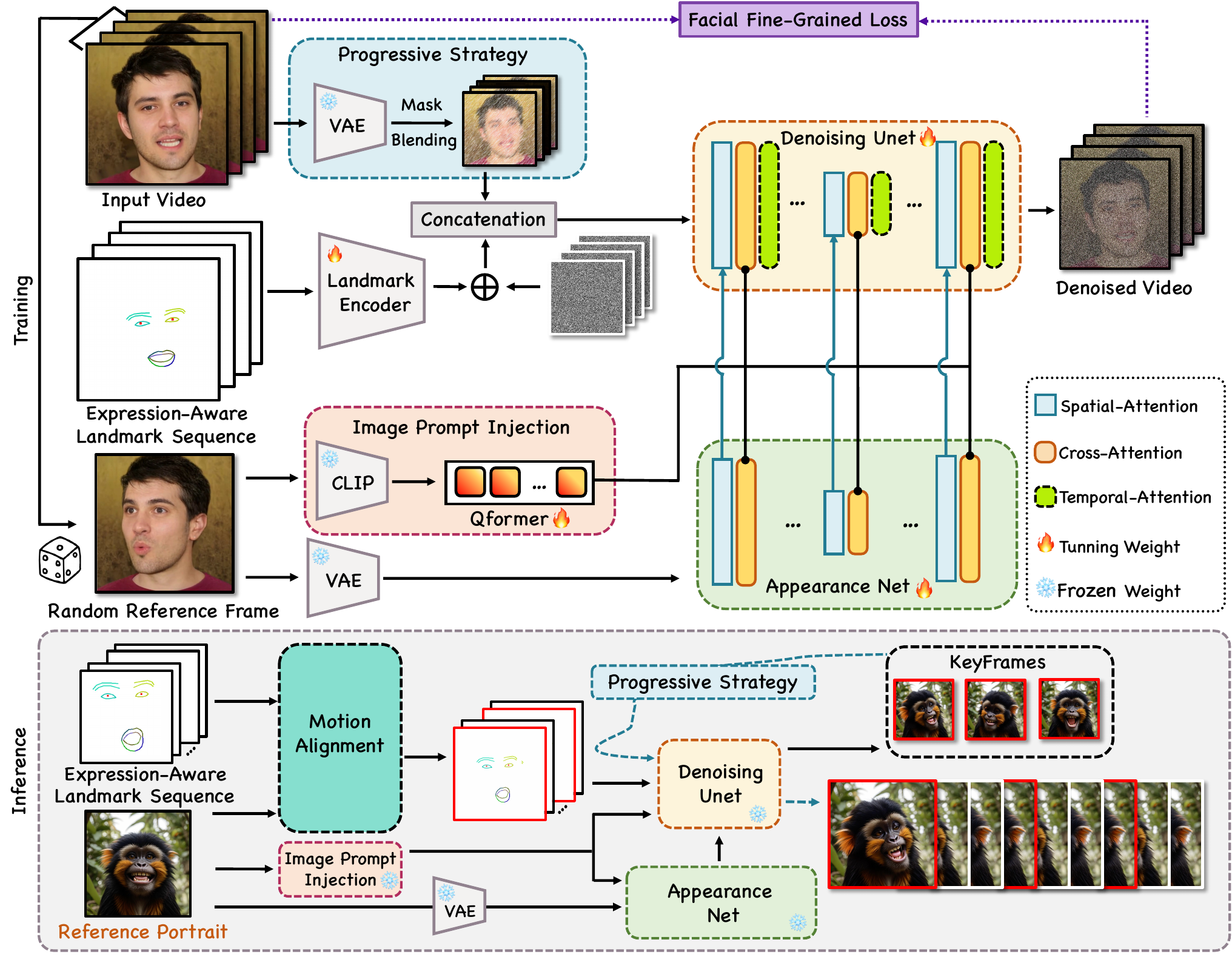} 
  \caption{\textbf{The overview of Follow-Your-Emoji}. We extract the features of our expression-aware landmark sequence with a landmark encoder and fuse these features with multi-frame noise first, then we utilize the progressive strategy to mask the frame of the input latent sequence randomly.  Finally, we concatenate this latent sequence with the fused multi-frame noise and feed it to the Denoising UNet to conduct the denoising process for video generation. The appearance net and image prompt injection module help our model preserve the identity of the reference portrait, and the temporal attention maintains the temporal consistency. During training, the facial fine-grinded loss guides the Unet to pay more attention to the facial and expression generation. During inference, following AniPortrait~\cite{wei2024aniportrait}, we align the target landmark with the reference portrait with the motion alignment module. Then, we first generate the keyframes and utilize the progressive strategy to predict long videos. }
  \vspace{-0.18in}
  \label{fig:framework}
\end{figure*}

\section{Method}
\label{sec:Method}
The pipeline of our method is shown in Fig.~\ref{fig:framework}. Given an input video clip, we randomly select a frame $\mathcal{I}_{0}$ as the reference portrait image. Then, we extract the motion sequences $\{L_{1}, L_{2}, L_{3}, ..., L_{N} \}$ (expression-aware landmarks) from the input video. The purpose of our method is to transfer the expression of the landmark sequences to $\mathcal{I}_{0}$. Even for reference portraits of uncommon styles (i.e., cartoon, sculpture, and animal), we hope our method can still predict good results.

 \begin{figure}[t]
  \centering
  \includegraphics[width=\linewidth]{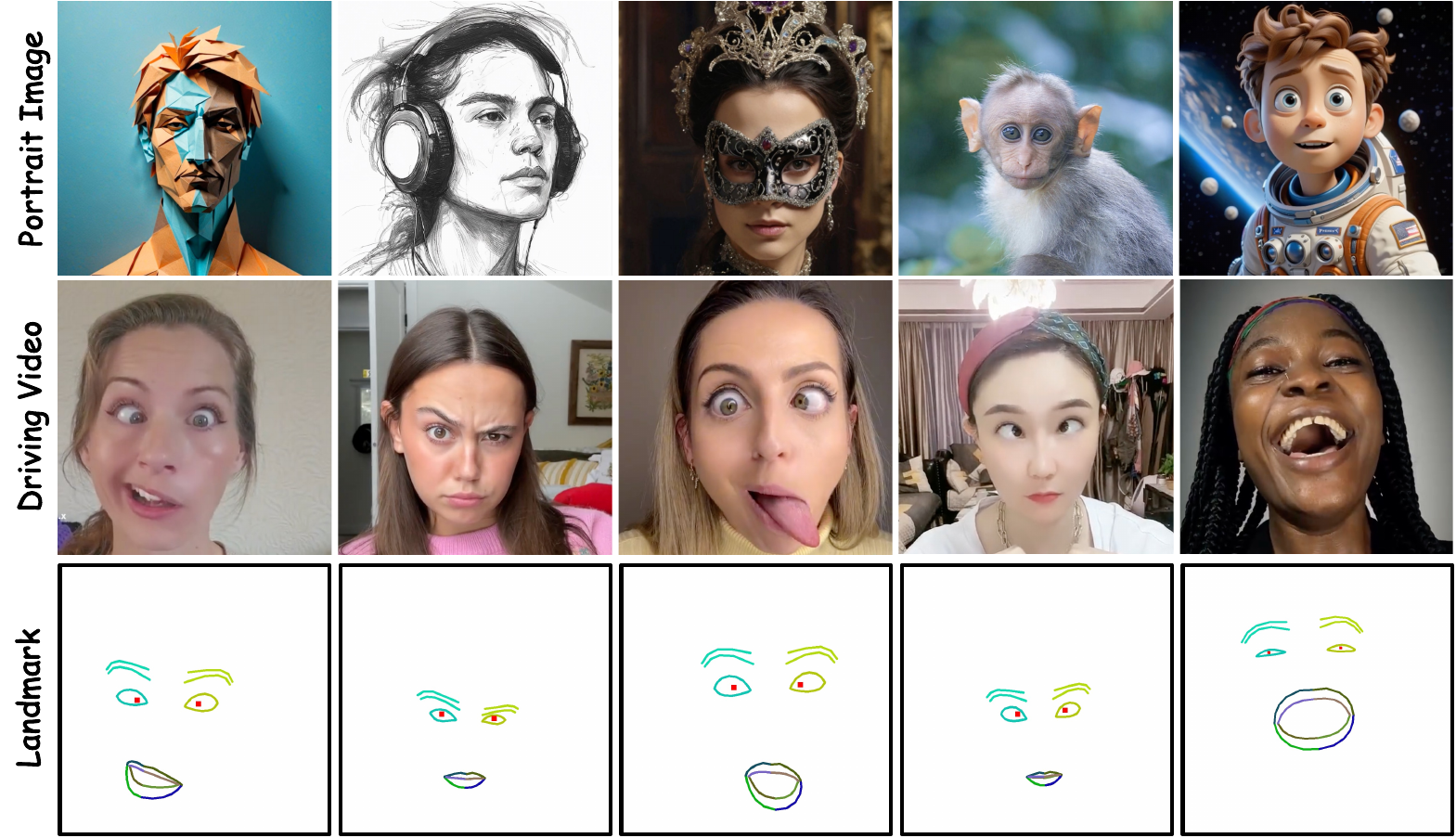} 
  \caption{\textbf{Examples of the EmojiBench with high expression diversity}, exaggeration, and various visual styles in portrait images.
  }
  \vspace{-0.18in}
  \label{fig:emojibench}
\end{figure}

 We follow the recent diffusion-based portrait animation methods in our framework and utilize both the appearance net and temporal attention. For the control motions injection, we add the features of our expression-aware landmarks to UNet directly. These features are extracted with a landmark encoder. Moreover, similar to StyleCrafter~\cite{liu2023stylecrafter}, we encode the reference image $\mathcal{I}_{0}$ to image token using pre-trained CLIP image encoder, then the 4-layers Qformer~\cite{zhang2023vision} is employed to fuse all image token. In the next,  we first discuss the motion representation and present our expression-aware landmark in Sec.~\ref{sec:Expression Aware Landmark}. Then, we introduce the facial fine-grained loss in  Sec.~\ref{sec:Facial Fine-grained loss}. Finally, for long-term animation, we describe the progressive strategy in Sec.~\ref{sec:Progressive Strategy}.

\subsection{Expression-Aware Landmark} 
\label{sec:Expression Aware Landmark}

Motion representation of facial expressions is essential for portrait animation. Accurate and precise motion representation enables conveying the nuances of human emotion and expression, thereby enhancing the overall realism and impact of the animated portrait. Recent diffusion-based methods always directly utilize the portrait image sequences providing the driving motion~\cite{xie2024x} or the 2D landmarks as the motion representation for training. However, during the inference process, 2D landmarks cannot ensure alignment between the target expression and the reference portrait. This misalignment will lead to inaccurate generated expressions and potential leakage of the identity information. Directly using the portrait image providing the driving motion can solve this problem, but it is necessary to ensure that the person in the motion sequence is different from the reference portrait during the training process, which requires another portrait animation method for identity conversion. This conversion process will damage the accuracy of the expressions, and the portrait animation method can not transfer the identity of the uncommon portrait (i.e., turning a dog into a human). 

To address the above problems, we introduce the expression-aware landmark, a new motion representation for portrait animation. Specifically, we utilize MediaPipe to extract the 3D keypoints of the portrait from the motion video. We then project these keypoints to obtain the 2D landmark. During the projection process, we discard the facial contour while retaining only the facial features. We find this operation can help the model focus on subtle motion generation and avoid the inaccuracy of facial contour with large expression changing, as shown in Fig.~\ref{fig:ablation_lmk}.  Moreover, to capture the motion of the portrait's irises, we calculate the related position of the irises in the eye sockets of 3D keypoints and maintain such a relationship after projection. In the end, since our expression-aware landmark is built on the 3D keypoints, we can align the target landmark sequence to the reference portrait in the canonical space of MediaPipe naturally, and we denote this process as motion alignment in the inference step as shown in Fig.~\ref{fig:framework}.

\subsection{ Facial Fine-Grained Loss}
\label{sec:Facial Fine-grained loss}
For the portrait animation task, we hope the diffusion model focuses on expression generation and identity preservation. However, the diffusion model's original training objective $\mathcal{L}_{LDM}$ is to learn the content of all regions of the target image, which has no specific constraints for learning the facial content during the training process. Therefore,  we propose the facial fine-grained (FFG) loss to modify the $\mathcal{L}_{LDM} $ and make the model pay more attention to the content of facial and expression regions.

\begin{figure}
  \centering
  \includegraphics[width=\linewidth]{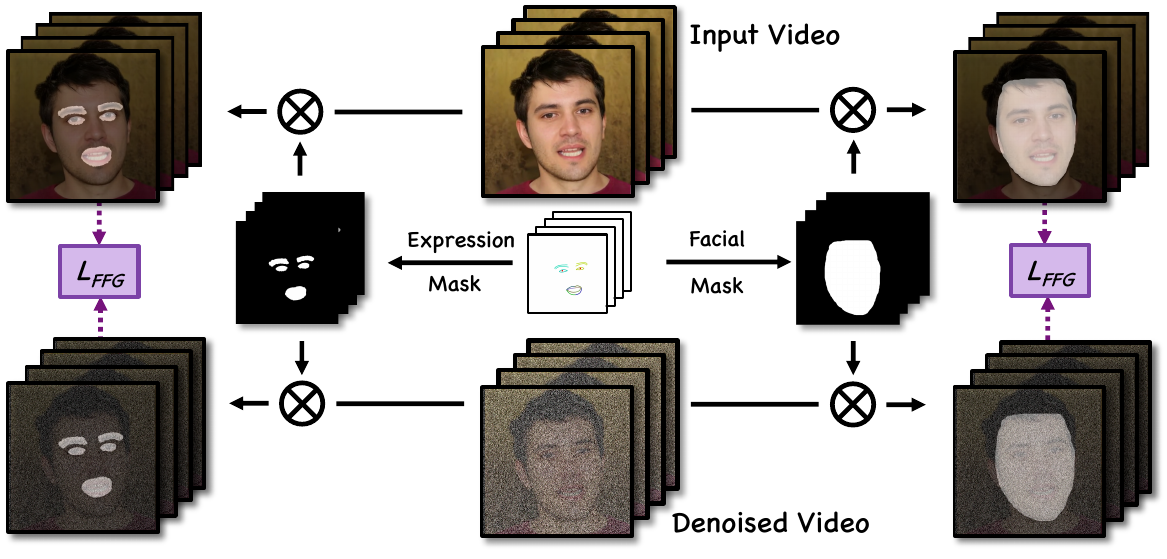} 
  \caption{\textbf{The detail of our facial fine-grained loss.} We extract the facial mask and expression mask with our landmark first. Then, we calculate the denoising loss $\mathcal{L}_{FFG} $  in these masked regions.  }
  \vspace{-0.18in}
    \label{fig:ffgloss}

\end{figure}

As shown in Fig.~\ref{fig:ffgloss}, we need to get two types of masks to capture the expression and facial regions to calculate the FFG loss.  For the expression mask ${\mathcal{M}_e} $,  we dilate each point of our expression-aware landmark and set these dilation regions as the expression mask. For the facial mask ${\mathcal{M}_f} $, we project the MediaPipe 3D facial counter's keypoints and connect these projected points to get the facial masks. Finally, these two masks split the FFG loss into expression and facial aspects, respectively. Formally, the loss function can be written as below:

\begin{equation}
\begin{aligned}
\mathcal{L}_{FFG}={\mathbb{E}}\left[\left\|\mathcal{M}_e \cdot \left(z-\hat{z} \right)+\mathcal{M}_f \cdot \left(z-\hat{z}\right)\right\|^2\right] \\
\label{eq:ffg_loss}
\end{aligned}
\end{equation} where $\hat{z}$ is the prediction latent embedding obtained by decoding the $\epsilon_{\theta}$. With our FFG loss, our method demonstrates better performance in both identity preservation and expression generation, as shown in Fig.~\ref{fig:ablation_ffgloss}. Finally, our total loss can be written as:
\begin{equation}
    \begin{aligned}
        \mathcal{L} =\mathcal{L}_{LDM}+\mathcal{L}_{FFG}
    \end{aligned}
\end{equation}

\subsection{Progressive Strategy for Long-Term Animation}
\label{sec:Progressive Strategy}
With the advancement of technology and increasing user demands, long-term animation has become increasingly important in practical applications. Despite training on video clips, previous approaches~\cite{magicpose, xu2023magicanimate, zhu2024champ, hu2023animate} have also attempted to generate long videos during testing. They always synthesize several overlapping video clips and merge them using Gaussian smoothing. However, we observe that this trick leads to the degradation of temporal consistency. 

To alleviate the above issues, a progressive strategy is proposed to generate long-term animation from coarse to fine.  Intuitively, to generate the long-term animation in the inference step, we hope to generate keyframes first and then use these keyframes to generate the long-term animation with interpolation operation. To simulate this process, apart from the first and last latent frames, we cover the other input video latent frames first. Then, we concatenated this covered video latent with original UNet inputs to do the denoising process. With this strategy, we can set the first and last frames as keyframes in the inference step and help our model generate long-term animation. Meanwhile, we also cover each latent frame of the input video with a probability of 0.5, which helps our model generate the keyframes's content in the first inference step since we need to cover all latent frames in this inference step. During the training process, we switch between these two covering strategies with a probability of 0.5.

\section{Experiment}

\begin{table*}[t!]
\centering
\caption{\textbf{Quantitative comparisons with SOTA baselines.} We evaluate our framework both self and cross reenactments on $256 \times 256$ test images. }
\vspace{-2mm}
\label{tab:quant_rec}
\scalebox{0.8}
{\begin{tabular}{lcccc|ccc|ccc}
\toprule
\multirow{2}{*}{Method}  & \multicolumn{4}{c}{\textbf{Self Reenactment}} & \multicolumn{3}{c}{\textbf{Cross Reenactment}} & \multicolumn{3}{c}{\textbf{User Study}} \\ \cmidrule(lr){2-5} \cmidrule(lr){6-8} \cmidrule(lr){9-11}
 & \textbf{L1}\ $\downarrow$  & \textbf{SSIM}\ $\uparrow$  & \textbf{LPIPS}\ $\downarrow$  & \textbf{FVD}\ $\downarrow$ & \textbf{ID Similarity}\ $\uparrow$ &\textbf{Image Quality}\ $\uparrow$ & \textbf{Expression}\ $\downarrow$ & \textbf{Motion}\ $\downarrow$ & \textbf{Identity}\ $\downarrow$ & \textbf{Overall}\ $\downarrow$\\
\midrule
Face Vid2vid~\citep{wang2021facevid2vid}   &0.043    & 0.792 & 0.258 & 232.1 & 0.614 & 37.295 & 5.42 &6.47 & 7.79 & 5.93  \\
DaGAN~\citep{hong2022depth}  &0.057    & 0.711 & 0.301 & 267.4 & 0.271 & 36.901 & 1.87 &7.18 & 6.83 & 7.26  \\
TPS~\citep{TPS}   &0.037    & 0.823 & 0.211 & 196.3 & 0.481 & 35.172 & 14.83 &4.94 & 4.57 & 4.81 \\
MCNet~\citep{MCNet}   &0.032    & 0.835 & 0.198 & 211.7 & 0.373 & 32.154 & 17.25 &5.53 & 5.11 & 5.89 \\
FADM~\cite{zeng2023face}  &0.048    & 0.693 & 0.274 & 191.4 & 0.672 & 32.493 & 2.14 &3.14 & 3.73 & 3.76 \\
MagicDance~\citep{magicpose}   &0.046    & 0.749 & 0.174 & 156.2 & 0.679 & 56.804 & 34.21 &2.65 & 2.56 & 2.74 \\
\midrule
Ours  &\textbf{0.029}    & \textbf{0.849} & \textbf{0.136} & \textbf{96.8}  & \textbf{0.702} & \textbf{66.287} & \textbf{39.16} &\textbf{1.27} & \textbf{1.48} & \textbf{1.78} \\ 
\bottomrule
\end{tabular}}
\vspace{-3mm}
\end{table*}

\begin{figure*}[t]
  \centering
  \includegraphics[width=1\linewidth]{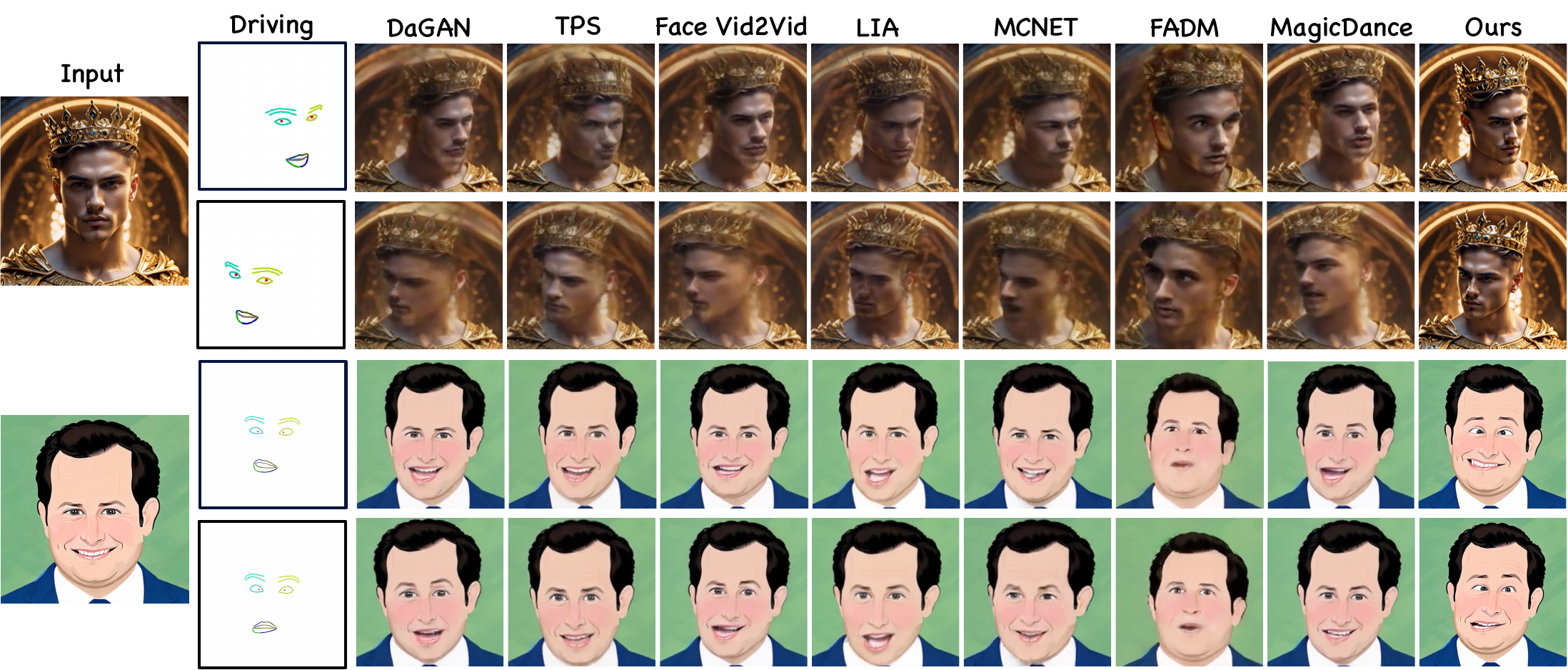} 
  \caption{\textbf{The qualitative comparisons with existing methods.} Given a reference portrait image and expression-aware landmarks, our approach demonstrates superior performance in capturing detailed facial expressions and maintaining the original identity of the characters compared to previous methods. More results are available in the supplementary material.}

  \label{fig:compare}
\end{figure*}

\subsection{Implementation Details}
We train our model on HDTF\cite{zhang2021flow}, VFHQ\cite{xie2022vfhq}, and our collected dataset jointly, which includes monocular camera recordings of 18 expressions and 20-minute real-human video from 115 subjects in both indoor and outdoor scenes.
The training stage consists of two stages, in the initial training stage, we sample individual video frames and perform resizing and center-cropping to achieve a resolution of $512 \times 512$. We fine-tune the model for 30,000 steps using a batch size of 32. In the subsequent training stage, we focus on training the temporal layer for 10,000 steps using 16-frame video sequences with a batch size of 32.
The learning rate is $1 \times 10^{-5}$ in two stages. The temporal attention layers are initialized with AnimateDiff~\cite{guo2023animatediff} similar to the AnimeAnyone. 
The frozen image autoencoder is applied to project each video frame into latent space. We optimize overall framework using Adam~\cite{loshchilov2017decoupled} on on 32 NVIDIA A800 GPUs. During inference, we utilize DDIM sampler~\cite{song2020denoising} and set the scale of classifier-free guidance~\cite{ho2022classifier} to 3.5 in our experiment.

\begin{table*}[t!]
\centering
\caption{ \textbf{Quantitative results of ablation study.} All metrics are evaluated on $256 \times 256$ test images. ${\uparrow}$ indicates higher is better. ${\downarrow}$ indicates lower is better.}
\vspace{-2mm}
\scalebox{0.9}
{
\begin{tabular}{lcccc|ccc}
\toprule
\multirow{2}{*}{Method}  & \multicolumn{4}{c}{\textbf{Self Reenactment}} & \multicolumn{3}{c}{\textbf{Cross Reenactment}} \\ \cmidrule(lr){2-5} \cmidrule(lr){6-8}
 & \textbf{L1}\ $\downarrow$  & \textbf{SSIM}\ $\uparrow$  & \textbf{LPIPS}\ $\downarrow$  & \textbf{FVD}\ $\downarrow$ & \textbf{ID Similarity}\ $\uparrow$ &\textbf{Image Quality}\ $\uparrow$ & \textbf{Expression}\ $\downarrow$ \\
\midrule
FFG Loss (w/o Expression Mask)  &0.037 & 0.702 & 0.159 & 147.4 & 0.576 & 53.792 & 26.87 \\
FFG Loss (w/o Identity Mask)  &0.036 & 0.721 & 0.157 & 149.3 & 0.548 & 50.992 & 34.21 \\
w/o Progressive Strategy  &0.035 & 0.718 & 0.141 & 138.8 & 0.632 & 52.108 & 35.32 \\
\midrule
2D Landmarks  &0.039 & 0.715 & 0.166 & 144.1 & 0.521 & 50.829 & 17.45 \\
w Facial Contour points  &0.034 & 0.784 & 0.153 & 128.5 & 0.627 & 59.781 & 38.71 \\
w/o Pupil points   &0.035 & 0.762 & 0.147 & 103.2 & 0.648 & 61.436 & 33.58 \\
\midrule
Ours  &\textbf{0.029} & \textbf{0.849} & \textbf{0.136} & \textbf{96.8}  & \textbf{0.702} & \textbf{66.287} & \textbf{39.16} \\ 
\bottomrule
\end{tabular}
}
\label{tab:ablation}
\vspace{-3mm}
\end{table*}

\subsection{EmojiBench}
We introduce EmojiBench, a new benchmark to evaluate the model's ability to animate freestyle portraits.  Specifically, we collect 410 portraits from different domains, including cartoon style, real-human style, and even animals.
These portrait cases are generated from 20 different personalized text-to-image models. We also provide 20 animal portraits, whose landmarks are able to be detected by Mediapipe~\cite{mediapipe}. The EmojiBench contains 45 videos of driving human heads collected from the internet. Each video is approximately 5 seconds long with 150 frames. The expressions of EmojiBench include a diverse range of head motion and facial expressions (\textit{e.g.}, frowning, crossed eyes, and pouting). Such a benchmark with various styles would be beneficial for the development of the community. 


\subsection{Comparison with baselines}
\subsubsection{Qualitative results.} 
\begin{figure*}
  \centering
  \includegraphics[width=\linewidth]{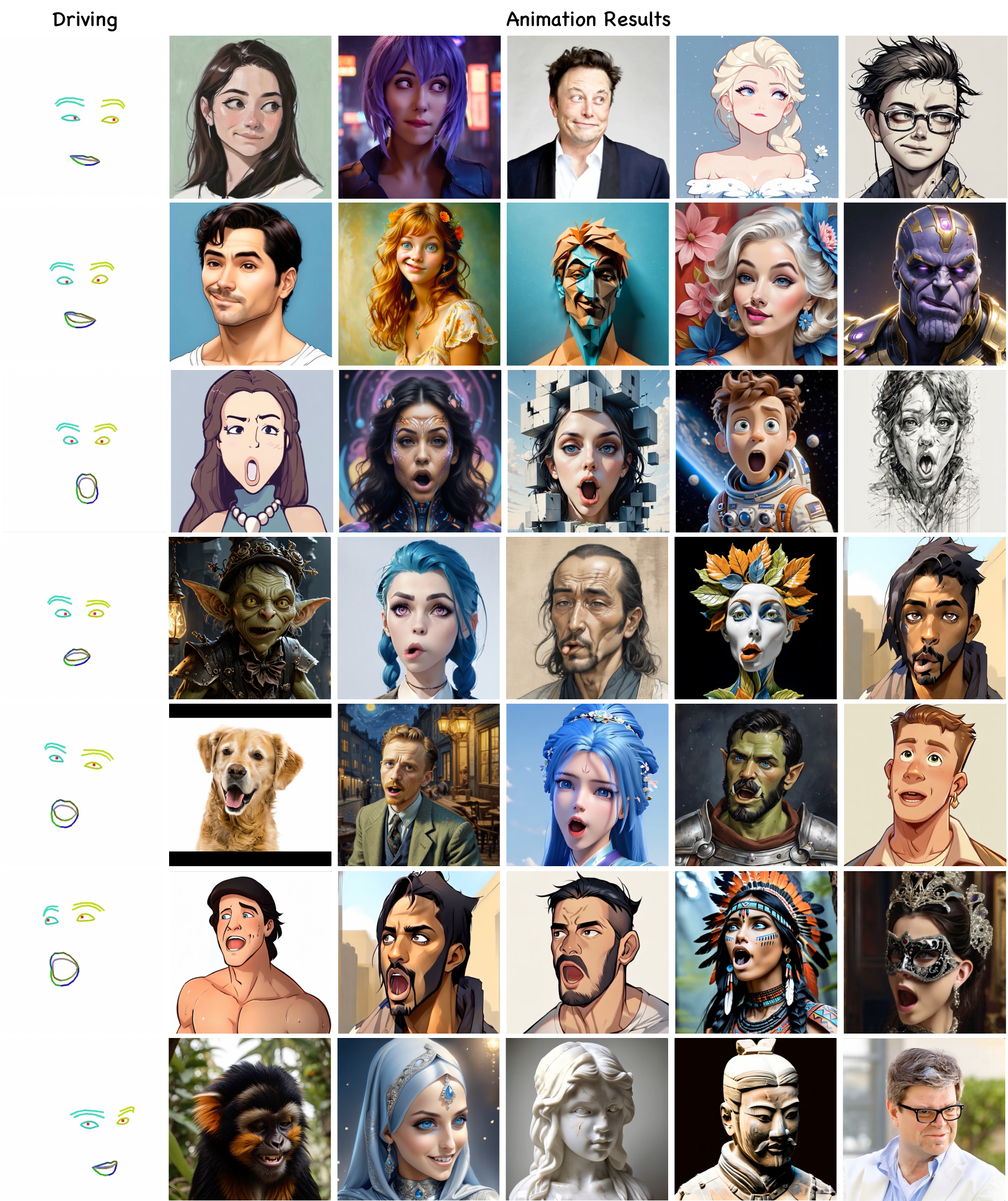} 
  \caption{More portrait animation results.  }
    \label{fig:appendix_1}

\end{figure*}

\begin{figure*}
  \centering
  \includegraphics[width=\linewidth]{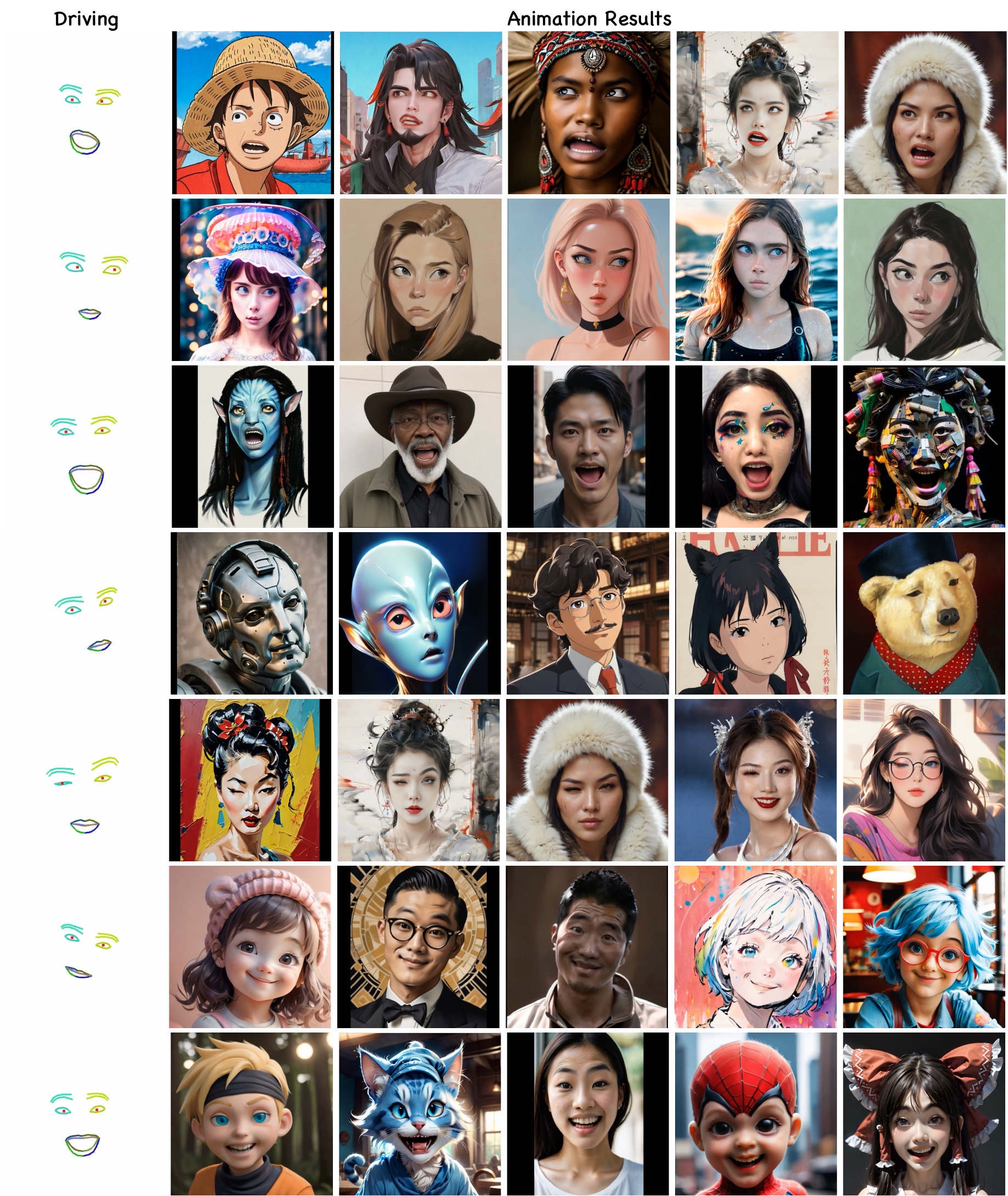} 
  \caption{More portrait animation results.  }
    \label{fig:appendix_2}

\end{figure*}

We compare our approach with previous portrait animation methods, including state-of-the-art GAN-based approaches Face Vid2vid~\cite{wang2021facevid2vid},  DaGAN~\cite{hong2022depth}, MCNet~\cite{MCNet}, TPS~\cite{TPS}.
Additionally, we also compare our method with concurrent diffusion-based methods like FADM~\cite{zeng2023face} and MagicDance~\cite{magicpose}. MegaPortraits~\cite{megaPortriat} and X-portrait~\cite{xie2024x} are excluded from our comparisons as no public release exists. 
The results are shown in Fig.~\ref{fig:compare}. We find the GAN-based method easily suffers from obvious artifacts, especially when changing the pose of the head with a large angle (i.e., see the generation result of the first character). Moreover, they can not rebuild the subtle expression for the reference portraits of uncommon style well (i.e., the movement of pupils in the second character). The diffusion-based methods MagicDance~\cite{magicpose} and  FADM~\cite{zeng2023face} perform better in expression transfer, but they still can not preserve the identity of reference portraits during animation. In contrast, our approach exhibits superior ability in handling large pose changing, subtle expressing generation, and identity preservation for uncommon style portraits. Please see more animation results in Fig.~\ref{fig:appendix_1} and Fig.~\ref{fig:appendix_2}.
\begin{figure}[t]
\centering
\includegraphics[width=1\linewidth]{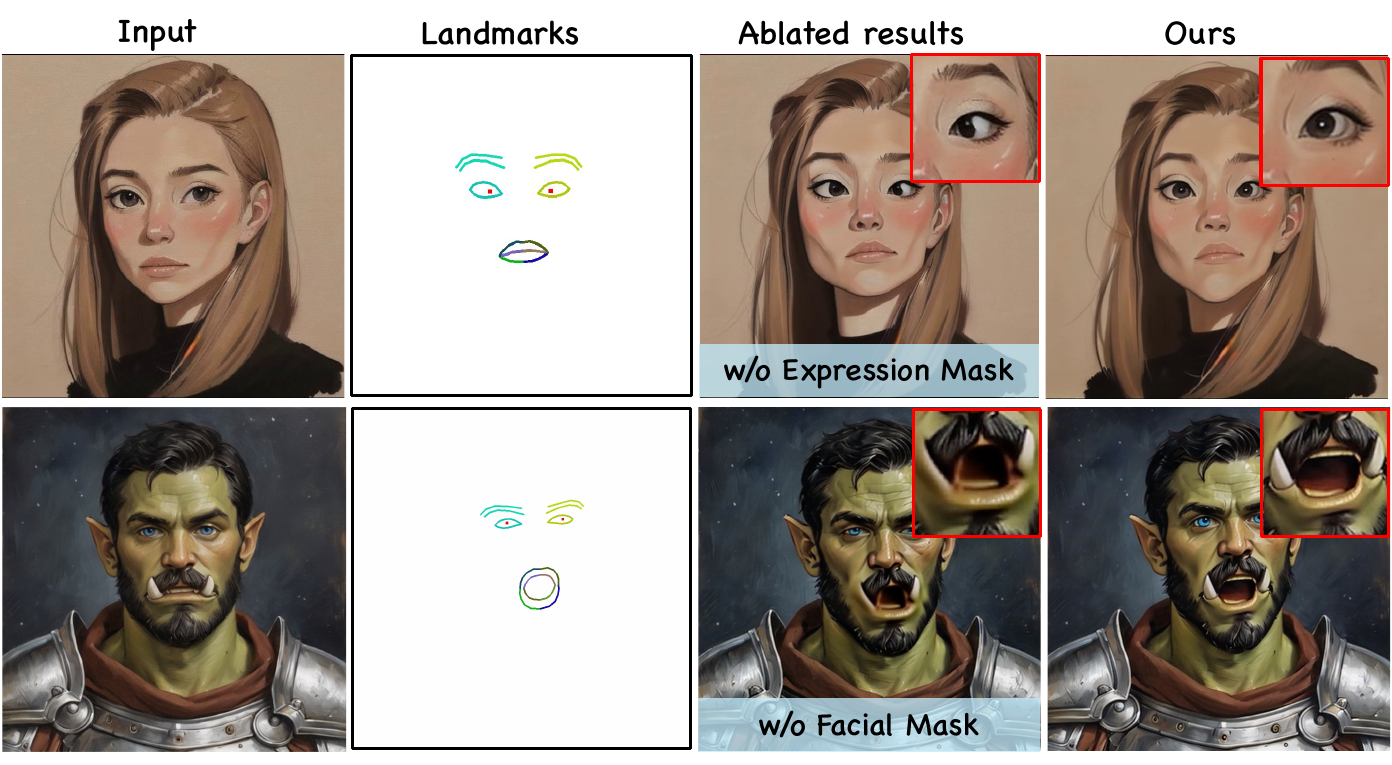} 
\caption{\textbf{The effectiveness of facial fine-grained loss.} We analyze the performance of expression and facial aspects of FFG loss, respectively. }
\label{fig:ablation_ffgloss}
\end{figure}

\subsubsection{Quantitative results.}
We compare our method with state-of-the-art portrait animation on our EmojiBench quantitatively and the results are shown in Tab.~\ref{tab:quant_rec}.  Due to the limited resolution of most previous works, all measurements are performed in 64 frames at a resolution of $256 \times 256$. All evaluation metrics used are as follows: (a) \textbf{Self Reenactment}: For quantitative assessment of image-level quality, we report the four metrics, L1 error, SSIM~\cite{wang2004image}, LPIPS~\cite{zhang2018unreasonable}, and FVD~\cite{unterthiner2018towards}.  For each video in EmojiBench, the first frame is employed as the reference image to generate the facial expression sequences. We leverage subsequent frames to serve as both the driving image and the ground truth. 
(b) \textbf{Cross Reenactment}: 
We evaluate cross reenactment on four metrics: identity similarity, image quality,  expression landmark accuracy, and user study, respectively.
(1) \textit{Identity similarity}: the ArcFace score~\cite{deng2019arcface} is applied to measure identity preservation.  We calculate cosine similarity between source and generated images.
(2) \textit{Image quality assessment}: We follow~\cite{xie2024x} to utilize the HyperIQA~\cite{zhang2023liqe} for image quality assessment. 
(3) \textit{Landmark accuracy}: To evaluate the pose accuracy of the generated video, we regard the input facial landmark sequences as ground truth and evaluate the average precision of the facial landmark sequences. 
(c) \textbf{User Study}: we perform the user study on cross reenactment with three aspects. (1) \textit{Expression:} Evaluating the quality of generated expression. (2) \textit{Identity:} Measuring the identity similarity between the generated frame images and input reference portrait image. (3)\textit{Overall:} Evaluating the overall quality of the generated videos. We randomly selected 45 cases and asked 30 volunteers to rank different methods in these three aspects. According to the results presented in Table~\ref{tab:quant_rec}, our approach demonstrates superior performance across seven metrics of self/cross reenactment. In terms of the user study, our approach outperforms previous baselines in terms of temporal coherence and identity preservation, while also exhibiting superior motion quality.

\subsection{Ablation Study}
In the subsequent section, we will analyze the effectiveness of expression-aware landmark and facial fine-grained loss. As for progressive strategy for long-term animation, we provide more discussion in supplementary materials.

\begin{figure}[t]
  \centering
  \includegraphics[width=1\linewidth]{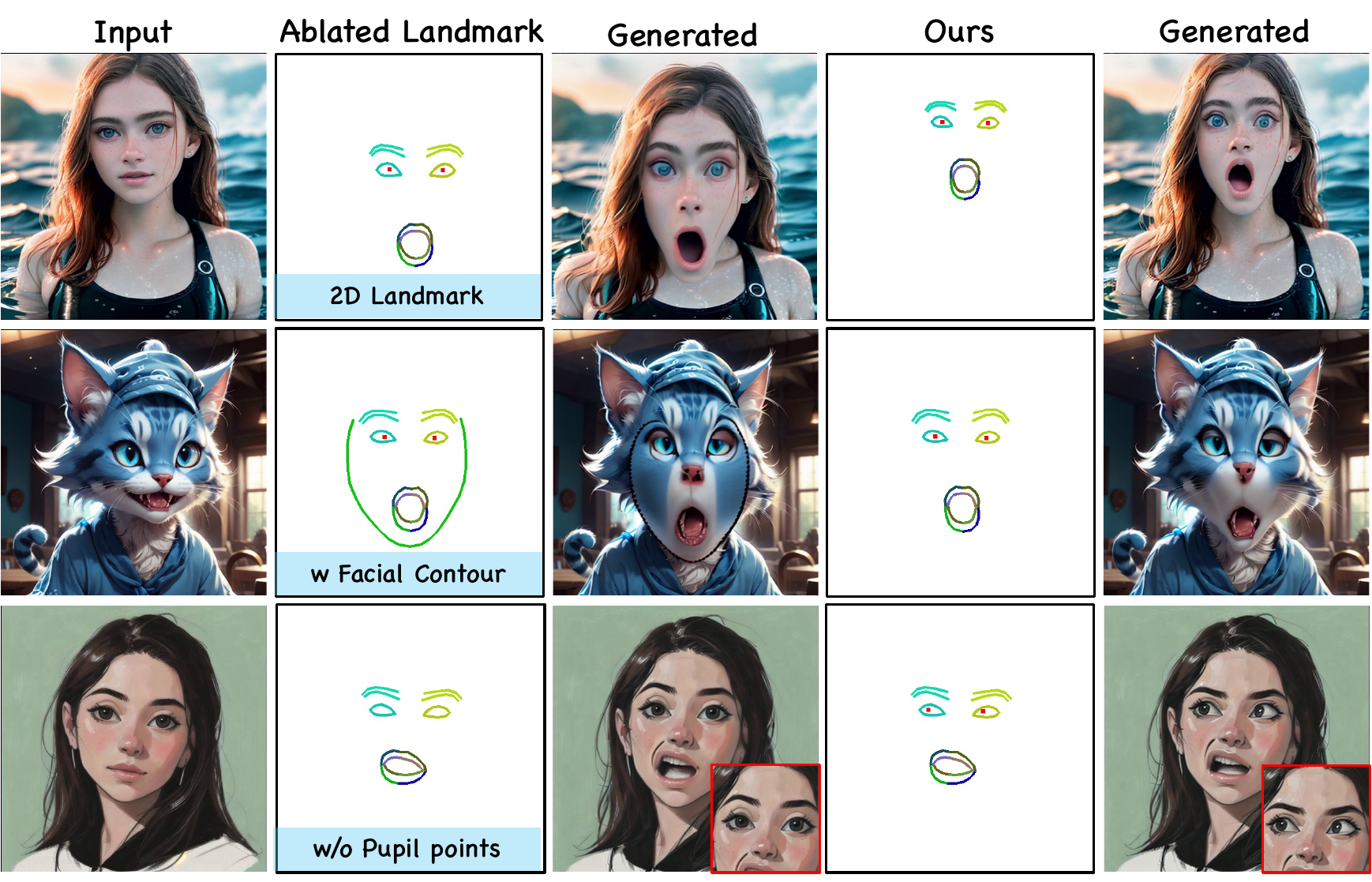} 
  \caption{\textbf{The effectiveness of expression-aware landmark.} We compare the results when different landmarks is used to guide the portrait animation. }
  \label{fig:ablation_lmk}
\end{figure}
\noindent\textbf{Effectiveness of Expression-Aware Landmark.} 
To prove the effectiveness of our expression-aware landmark, we change our motion representation to the 2D landmark, expression-aware landmark with the facial counter, and expression-aware landmark without pupil points to generate the video, respectively. The visual results are shown in Fig.~\ref{fig:ablation_lmk}. 2D landmark has a challenge in handling the alignment of the facial bounding box between target landmarks and reference portrait images, as presented in the 1st row. The expression-aware landmark with the facial counter fails to maintain the identity of  portrait images in non-human styles. This is because the current open-source landmark detector makes it hard to predict the facial counter of any style portrait. Finally, we also show the result produced with expression-aware landmarks without pupil points. Due to the lack of motion signals of pupil points,  it is difficult to generate lively expressions with pupil motion. In contrast, our full model demonstrates better performance. The corresponding numerical evaluation is shown in Tab.~\ref{tab:ablation}.

\noindent\textbf{Effectiveness of Facial Fine-Grained Loss.} To analyze the performance of FFG loss, we discarded the expression and facial aspects of FFG loss separately to do the experiment. Without facial aspects of FFG loss, we find our method reduces the ability to protect identity information and detail appearance of the input portrait (i.e., teeth disappeared in the second row of Fig.~\ref{fig:ablation_ffgloss}). Meanwhile, when we abandon the expression aspects of FFG loss, our method can not capture the subtle expression changing well (i.e., inaccurate pupil movement in the first row of Fig.~\ref{fig:ablation_ffgloss}). The corresponding numerical evaluation is shown in Tab.~\ref{tab:ablation}.

\section{Conclusion}
We introduce Follow-Your-Emoji, a novel diffusion-based framework for freestyle portrait animation. Incorporating with the expression-aware landmark, our method shows high performance in subtle and exaggerated facial expression generation. Meanwhile,  we propose a facial fine-grained loss to constrain the diffusion model focus on expression generation and identity preservation. To train our model, we introduce a new expression training dataset with 18 exaggerated expressions and 20-minute real-human videos from 115 subjects. Then, we introduce the progressive strategy for stable long-term animation. Finally, to address the lack of benchmark in portrait animation, we build the EmojiBench, a comprehensive benchmark to evaluate our method, the impressive performance of our model on generalized reference portraits and driving motions serves as validation of its effectiveness.

\paragraph{Acknowledgments.}
We thank Jiaxi Feng, Yabo Zhang for their helpful comments. 
This project was supported by the National Key R\&D Program of China under grant number 2022ZD0161501.

\clearpage

\clearpage

{
    \small
    \bibliographystyle{ieeenat_fullname}
    \bibliography{main}
}


\end{document}